\documentclass{article}

\usepackage{arxiv}
\usepackage{amsmath}
\usepackage[utf8]{inputenc} 
\usepackage[T1]{fontenc}    
\usepackage{hyperref}       
\usepackage{url}            
\usepackage{booktabs}       
\usepackage{amsfonts}       
\usepackage{nicefrac}       
\usepackage{microtype}      
\usepackage{lipsum}		
\usepackage{graphicx}
\usepackage{natbib}
\usepackage{doi}
\usepackage{subcaption}

\title{Post-collision trajectory restoration for a single-track Ackerman vehicle using heuristic steering and tractive force functions}

\author{ \href{https://orcid.org/0009-0000-2924-3696}{\includegraphics[scale=0.06]{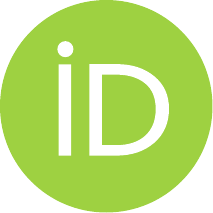}\hspace{1mm}Samsaptak Ghosh}\\
    Electrical Engineering Department\\
	Indian Institute of Technology Roorkee\\
	Roorkee, India \\
	\texttt{sghosh@ee.iitr.ac.in} \\
	\And
	\href{https://orcid.org/0000-0001-8500-7604}{\includegraphics[scale=0.06]{orcid.pdf}\hspace{1mm}M. Felix Orlando} \\
	Electrical Engineering Department\\
	Indian Institute of Technology Roorkee\\
	Roorkee, India \\
	\texttt{m.orlando@ee.iitr.ac.in} \\
    \And
	\href{https://orcid.org/0000-0001-7213-6693}{\includegraphics[scale=0.06]{orcid.pdf}\hspace{1mm}Sohom Chakrabarty} \\
	Electrical Engineering Department\\
	Indian Institute of Technology Roorkee\\
	Roorkee, India \\
	\texttt{sohom@ee.iitr.ac.in} \\
}

\renewcommand{\shorttitle}{Post-Collision trajectory restoration}

\hypersetup{
pdftitle={Post-collision trajectory restoration for a single-track Ackerman vehicle using heuristic steering and tractive force functions},
pdfsubject={q-bio.NC, q-bio.QM},
pdfauthor={Samsaptak Ghosh, M. Felix Orlando, Sohom Chakrabarty},
pdfkeywords={Post-collision trajectory restoration, heuristic steering and tractive force functions, Ackerman vehicle},
}

\begin{document}
\maketitle

\begin{abstract}
	Post-collision trajectory restoration is a safety-critical capability for autonomous vehicles, as impact-induced lateral motion and yaw transients can rapidly drive the vehicle away from the intended path. This paper proposes a structured heuristic recovery control law that jointly commands steering and tractive force for a generalized single-track Ackermann vehicle model. The formulation explicitly accounts for time-varying longitudinal velocity in the lateral--yaw dynamics and retains nonlinear steering-coupled interaction terms that are commonly simplified in the literature. Unlike approaches that assume constant longitudinal speed, the proposed design targets the transient post-impact regime where speed variations and nonlinear coupling significantly influence recovery. The method is evaluated in simulation on the proposed generalized single-track model and a standard 3DOF single-track reference model in MATLAB, demonstrating consistent post-collision restoration behavior across representative initial post-impact conditions.
\end{abstract}

\keywords{Generalized Ackermann Vehicle Model \and Post-collision trajectory \and Steering Function \and Tractive Force Function}

{A}{n} enormous technological leap has occurred with the advent of self-driving cars, and this development will profoundly alter future road travel. Improved traffic flow, energy efficiency, passenger comfort, and a decrease in human error-related road accidents are all possible outcomes of implementing autonomous driving technology. \cite{brembeck2020recent}. 
The rapid development of Artificial Intelligence-Assisted Compressed Sensing has resulted in the extensive use of Advanced Driver-Assistance System and other forms of autonomous driving technology in the automotive sector. Autonomous driving systems often incorporate functions such as path-tracking control, motion planning, and environmental sensing. These systems can steer, brake, and adjust the throttle with ease and safety, thanks to trajectory tracking control that incorporates both longitudinal control and lateral control \cite{{brembeck2020recent},{marzbani2019autonomous},{liu2018dynamic}}.
To solve the issue of lane change maneuvring, Liu et al.'s \cite{liu2018dynamic} work focuses on the dynamic modelling and control of high-speed autonomous vehicles.
Nonlinear vehicle dynamics, discrete target routes and transient disturbances are added to the experimental framework and the driver model is studied in \cite{{qin2022lateral}}. 
 An alternate multibody approach for tracked vehicles and a piecewise affine model for intelligent vehicles are presented in the literature. In order to explain how the wheel interacts with the ground, it includes a nonlinear contact force model that accounts for both hard and soft terrain \cite{sun2022piecewise}.
In \cite{pan2020validation}, a double-step semi-recursive multibody formulation for real-time simulation is proposed, and high-order numerical time-integration techniques are used to analyze the performance.
Rafatnia et al. \cite{rafatnia2022estimation} explained the reliable vehicle dynamics model with stability controller design using IMU/GNSS data fusion. In this work, the uncertainty in the model is considered for both the tyre forces by getting estimation from IMU data. 
Literature proposes various control strategies to help autonomous cars avoid dynamic obstacles, considering road width, vehicle geometry, as well as the driver's complex and unpredictable behavior \cite{zeng2023collision,xu2024modeling,yang2023collision}. In the literature, it assumes absolutely collision-free movement of vehicles, and thus falls short in addressing the important issue of unanticipated collisions and the responses thereof. As such, procedures for the vehicle's recovery in diverse collision scenarios are not included in present auto-driver algorithms \cite{{jahromi2015integrated}}.
\par These procedures are, however, important to be considered in the autonomous driving algorithms due to a number of reasons: (1) If an autonomous system fails, the vehicle model's collision responses and trajectory restoration after a collision must be studied. (2) Collisions can occur because of many manual-driving vehicles on the same track. (3) There are also environmental factors that can enhance the collision scenarios.
\par
In view of the above research gap, the paper contributes to the following: 
\begin{enumerate}
    \item A novel heuristic control function is proposed for the steering and tractive force inputs to restore the trajectory of a vehicle moving with a high longitudinal velocity and without using the brakes in post-collision scenarios.
    \item The control function is validated and compared for both the generalized Ackermann vehicle model \cite{ghosh2023collision} and the 3DOF MATLAB single-track vehicle model.
\end{enumerate}
The recently examined generalized Ackermann vehicle model \cite{ghosh2023collision} is considered due to its incorporation of nonlinear frictional drag forces, which account for dynamics due to variations in steering angle input, longitudinal velocity, and complex nonlinear interactions with the steering angle. The collisions are emulated by applying initial angular and lateral velocities at the vehicle model's center of gravity. Several collision situations are studied by combination of the initial angular and lateral velocities. In this paper, the post-collision recovery is considered for the original trajectory being a straight path only.
\par
The authors consider the following important assumptions, which define the scope of the work.
\begin{itemize}
    \item \textbf{Assumption 1:} The collision takes place laterally on the vehicle body. No head-on collision is considered in the present work.
    \item \textbf{Assumption 2:} Collision does not generate roll, pitch, or z-axis dynamics. In this work, the collision excites y-dynamics and yaw dynamics only, which are emulated by taking initial values of velocities in these directions.
    \item \textbf{Assumption 3:} Only a single collision is considered at t=0, which does not affect the structure of the vehicle model, i.e., the vehicle parameters remain unchanged post-collision.
\end{itemize}
\par The paper is organized as follows: The generalized vehicle model is presented from \cite{ghosh2023collision} in Section II, and the novel heuristic control function design in Section III, which is the original contribution of this paper. Section IV uses specific example collision scenarios to validate that the vehicle returns to its original trajectory due to the applied control function. Section V presents the concluding remarks.
\section{Vehicle Model}
\begin{figure*}[!htpb]
	\centerline{\includegraphics[width=\columnwidth]{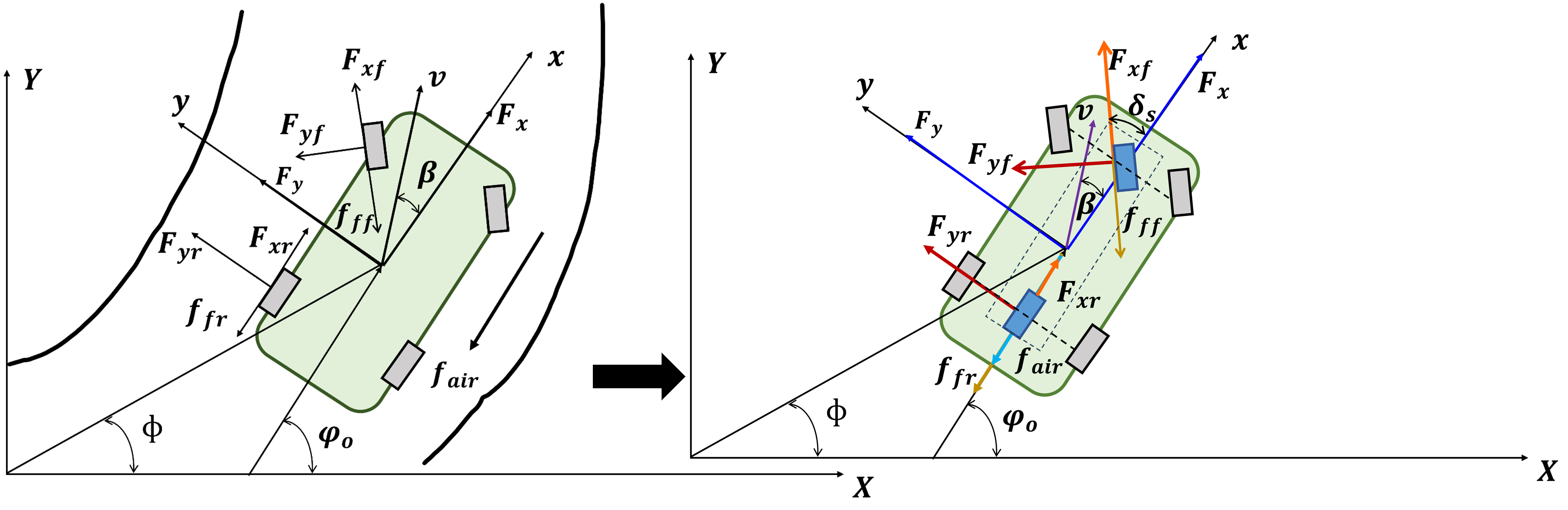}}
	\caption{Schematic diagram of the proposed generalized vehicle model showing the force components and velocity.
	}
	\label{Fig:1}
\end{figure*}
Figure ~\ref{Fig:1} shows the schematic of the forces and velocity components of the proposed single-track vehicle model in \cite{ghosh2023collision}. This schematic is used to determine the subsequent forces and moment acting on the vehicle body in the x, y, and z axes as,
\begin{equation}
\begin{aligned}
 F_x&= F_{xf}cos{\delta_s} + F_{xr} - F_{yf} sin{\delta_s} - f_{ff} cos{\delta_s} -f_{fr} -f_{air} \\ 
\end{aligned}
 \end{equation}
 \begin{equation}
 \begin{aligned}
       F_y&=F_{xf} sin{\delta_s} +F_{yf} cos{\delta_s} +F_{yr} - f_{ff} sin{\delta_s}\\   
 \end{aligned}
  \end{equation}
     \begin{equation}
         \begin{aligned}
                M_z&=l_f (F_{yf} cos(\delta_s) + F_{xf} sin(\delta_s) - f_{ff} sin(\delta_s))-l_r F_{yr}   
         \end{aligned}
     \end{equation}
where $l_f$ is the distance of the centre of gravity (C.G.) from the front wheel and $l_r$ denotes the distance of the rear wheel to the centre of gravity (C.G.).
The force components are expressed in the body frame ($B_f$), with respect to the global frame ($G_f$) as,
\begin{equation}
	\begin{aligned}
		{[F]}_{(B_f)}={{R^T}_{BG}}{[F]}_{(G_f)}= m{{R^T}_{BG}}{[\dot{V}]}_{(G_f)}
	\end{aligned}
\end{equation}
The simplification of the forces is taken from \cite{ghosh2023collision},
\begin{equation}
	\begin{bmatrix}
		F_x\\
		F_y\\
		0
	\end{bmatrix}=m\begin{bmatrix}
		\dot{v}_x-w_z v_y\\\dot{v}_y+w_z v_x\\0
	\end{bmatrix}
\end{equation}
where $v_x$, $v_y$ and $w_z$ signify the longitudinal, lateral and angular velocities of the vehicle with respect to the body frame (${B_f}$). Likewise, one may represent the moment in ${B_f}$ with respect to ${G_f}$ as {\cite{ghosh2023collision}},
\begin{equation}
	\begin{aligned}
		{[M]}_{(B_f)} &= {{R^T}_{BG}}{[M]}_{(G_f)}\\
		\begin{bmatrix}
			0\\
			0\\
			M_z
		\end{bmatrix} &= {{R^T}_{BG}}\begin{bmatrix}
			0\\
			0\\
			M_Z
		\end{bmatrix}
            = \begin{bmatrix}
			0\\
			0\\
			{I_z}{\dot{w_z}}
		\end{bmatrix}
		= \begin{bmatrix}
			0\\
			0\\
			{I_z}{\ddot{\psi_o}}
		\end{bmatrix}
	\end{aligned}
\end{equation}
\par
The final equation of the vehicle dynamics can be expressed as follows \cite{ghosh2023collision},
\begin{equation}
	\begin{aligned}
		\dot{v}_x &= \frac {F_{xt}}{m} (cos{\delta_s} + 1) -\frac {2C_{\alpha f}}{m} {\delta_s}sin{\delta_s} + \frac {2C_{\alpha f}}{m} \frac{v_y}{v_x} sin{\delta_s} 
		+ \frac {2C_{\alpha f}}{m} {l_f} \frac{w_z}{v_x} sin{\delta_s} \\ &- ({\mu}_0 + {{\mu}_1}{v_x}^2) g (cos{\delta_s} - 1)- \frac{1}{2} \frac{{\rho}{C_d}{A_F}}{m} ({v_x})^2 + w_z v_y
	\end{aligned}
\end{equation}
\begin{equation}
	\begin{aligned}
		\dot{v}_y &= \frac{F_{xt}}{m} sin{\delta_s} +\frac{2C_{{\alpha}f}}{m} {\delta_s}cos{\delta_s} -\frac{1}{m}{2C_{{\alpha}f}}cos{\delta_s}{\frac{v_y}{v_x}}-\frac{1}{m}{2C_{{\alpha}r}}  {\frac{v_y}{v_x}} - \frac{1}{m} {2C_{{\alpha}f}{l_f}} cos{\delta_s}\frac{w_z}{v_x}\\& +\frac{1}{m}{2C_{{\alpha}r}}{l_r} \frac{w_z}{v_x} - ({\mu}_0 + {{\mu}_1}{v_x}^2) gsin{\delta_s} - w_z v_x\\
	\end{aligned}
\end{equation}
\begin{equation}
	\begin{aligned}
		{\dot{w_z}}&=\frac{l_f}{I_z} 2C_{{\alpha}f} {\delta_s}cos{\delta_s} - \frac{1}{I_z} (2C_{{\alpha}f}{l_f}cos{\delta_s}- 2C_{{\alpha}r}{l_r}) \frac{v_y}{v_x} - \frac{1}{I_z} (2C_{{\alpha}f}{l_f}^2 cos{\delta_s} + 2C_{{\alpha}r}{l_r}^2) \frac{w_z}{v_x}\\
		& + \frac{l_f}{I_z} F_{xt} sin{\delta_s}- ({\mu}_0 + {{\mu}_1}{v_x}^2)\frac{mg{l_f}}{I_z} sin{\delta_s}  
	\end{aligned}
\end{equation}
\section{Force and Steering Control function}
This is the novel contribution of this paper wherein a heuristic steering and force control function is proposed to recover the vehicle's trajectory in post-collision scenarios.
\subsection{Steering Function}
The steering function control input ${\delta}_{s}$ can be defined as,
\begin{equation}
    {\delta}_{s}(t)={K_{dir}}[{\delta}_{1}(t)+{\delta}_{2}(t)]
\end{equation}
where ${\delta}_{1}(t)$ component is responsible for bringing the vehicle back towards the desired trajectory. The ${\delta}_{2}(t)$ component maintains the vehicle direction after coming to the desired trajectory. $K_{dir}$ is the constant that decides the vehicle's direction after collision. The ${\delta}_{1}(t)$ is designed as,
\begin{equation}
  {\delta}_{1}(t)=[{A_1}sin(w_1t-\phi_1)][u_c(t-{{\tau}_0})-u_c(t-{{\tau}_1})]
  \end{equation}
where $u_c(t-{\tau})$ denotes a unit step function, and is defined as $u_c(t-{\tau})=0$ for $t<\tau$ and $u_c(t-{\tau})=1$ for $t\geq\tau$. By considering $w_{1}=\frac{2\pi}{T_1}$ and phase delay $\phi_{1}=\frac{2\pi}{T_1}{\tau}_{0}$,
\begin{equation}
  {\delta}_{1}(t)=[{A_1}sin(\frac{2\pi}{T_1}t-{\frac{2\pi}{T_1}}{\tau}_0)][u_c(t-{{\tau}_0})-u_c(t-{{\tau}_1})]   
\end{equation}
where, ${\tau}_{0}$ is the steering function initialization time. It is clear that ${\delta}_{1}(t)$ is non-zero only between ${\tau}_{0}$ and ${\tau}_{1}$, elsewhere it is zero. The time period of ${\delta}_{1}(t)$ is chosen as ${T_1}=2({\tau}_{1}-{\tau}_{0})$. Then, ${\delta}_{1}(t)$ can be written as,
\begin{equation}
\begin{aligned}
    {\delta}_{1}(t)&=[{A_1}sin(\frac{\pi}{({\tau}_1-{\tau}_0)}t-{\frac{\pi}{({\tau}_1-{\tau}_0)}}{\tau}_0)][u_c(t-{{\tau}_0})-u_c(t-{{\tau}_1})]
    \end{aligned}
\end{equation}
Similarly, ${\delta}_{2}(t)$ is designed as,
\begin{equation}
    {\delta}_{2}(t)=[{A_2}sin(w_2t-\phi_2)][u_c(t-{{\tau}_2})-u_c(t-{{\tau}_3})]
\end{equation}
where ${\delta}_{2}(t)$ is non-zero between ${\tau}_{2}$ and ${\tau}_{3}$. By considering $w_{2}=\frac{2\pi}{T_2}$ and phase delay $\phi_{2}=\frac{2\pi}{T_2}{\tau}_{2}$,
\begin{equation}  
{\delta}_{2}(t)=[{A_2}sin(\frac{2\pi}{T_2}t-{\frac{2\pi}{T_2}}{\tau}_2)][u_c(t-{{\tau}_2})-u_c(t-{{\tau}_3})]
\end{equation}
The time period is chosen as ${T_2}=2({\tau}_{3}-{\tau}_{2})$. Then, ${\delta}_{2}(t)$ can be written as,
\begin{equation}
\begin{aligned}
     {\delta}_{2}(t)&=[{A_2}sin(\frac{\pi}{({\tau}_3-{\tau}_2)}t-{\frac{\pi}{({\tau}_3-{\tau}_2)}}{\tau}_2)][u_c(t-{{\tau}_2})-u_c(t-{{\tau}_3})]
    \end{aligned}
\end{equation}
So, the total steering function control input can be written as,
\begin{equation}
\begin{aligned}
    {\delta}_{s}(t)&=K_{dir}[[{A_1}sin(\frac{\pi}{({\tau}_1-{\tau}_0)}t-{\frac{\pi}{({\tau}_1-{\tau}_0)}}{\tau}_0)][u_c(t-{{\tau}_0})-u_c(t-{{\tau}_1})]\\&+[{A_2}sin(\frac{\pi}{({\tau}_3-{\tau}_2)}t-{\frac{\pi}{({\tau}_3-{\tau}_2)}}{\tau}_2)][u_c(t-{{\tau}_3})-u_c(t-{{\tau}_2})]]
    \end{aligned}
\end{equation}
Figure ~\ref{Fig:2}(a) provides an idea of the nature of this function.
\subsection{Tractive Force Function}
\begin{figure}[!htpb]
    \centering
    \begin{subfigure}{0.5\textwidth}
        \includegraphics[width=\linewidth]{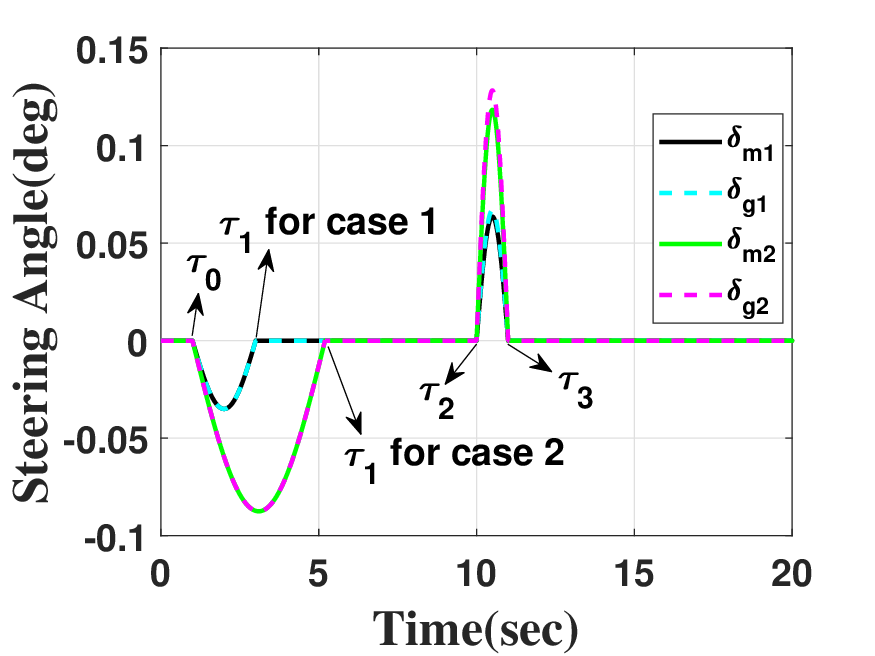}
        \caption{Steering angle ($\delta_s$)}
    \end{subfigure}\hfill
    \begin{subfigure}{0.5\textwidth}
        \includegraphics[width=\linewidth]{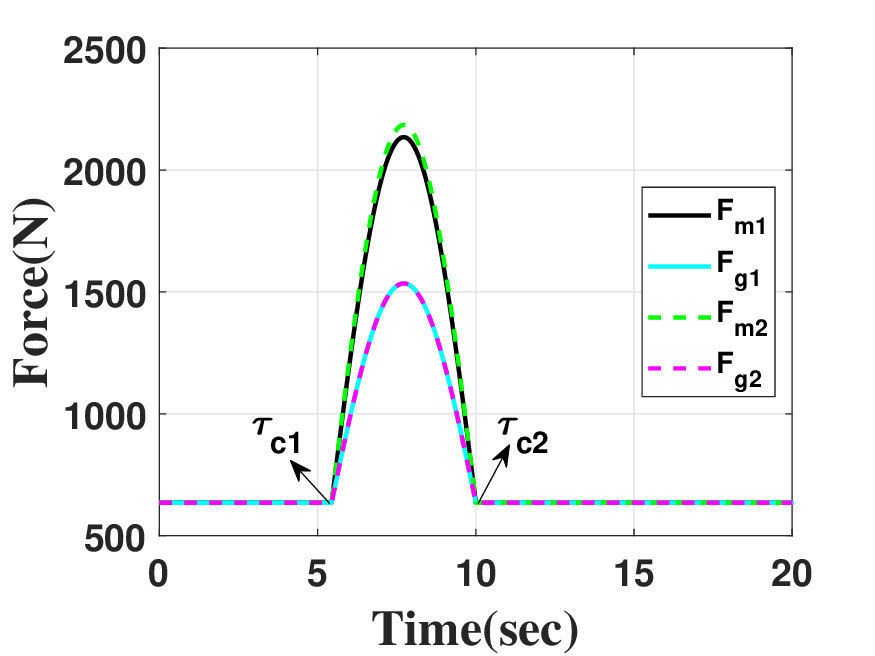}
        \caption{Tractive force ($F_{xt}$)}
    \end{subfigure}

    \vspace{1em}
    \caption{Case 1 and Case 2 collision scenario: (a) Steering angle ($\delta_s$), (b) Tractive force ($F_{xt}$)}
    \label{Fig:2}
\end{figure}

The tractive force function control input $F_{xt}$ can be defined as,
\begin{equation}
F_{xt}(t)=F_{i}(t)+F_c(t)
\end{equation}
where $F_{i}(t)$ is the initial force of the vehicle during the collision and $F_{c}(t)$ is the additional control force required to restore the trajectory after the collision. The control force is designed as,
\begin{equation}
F_c(t)=[{A_c}sin(w_ct-\phi_c)][u_c(t-{{\tau}_{c1}})-u_c(t-{{\tau}_{c2}})]
\end{equation}
By considering $w_{c}=\frac{2\pi}{T_c}$ and phase delay $\phi_{c}=\frac{2\pi}{T_c}{\tau}_{c1}$
\begin{equation}
F_c(t)=[{A_c}sin(\frac{2\pi}{T_c}t-\frac{2\pi}{T_c}{\tau}_{c1})][u_c(t-{{\tau}_{c1}})-u_c(t-{{\tau}_{c2}})]
\end{equation}
where $F_c(t)$ is non-zero between ${\tau}_{c1}$ and ${\tau}_{c2}$. The time period is chosen as ${T_c}=2({\tau}_{c2}-{\tau}_{c1})$. Then,
\begin{equation}
\begin{aligned}
F_c(t)&=[{A_c}sin(\frac{\pi}{({\tau}_{c2}-{\tau}_{c1})}t-\frac{\pi}{({\tau}_{c2}-{\tau}_{c1})}{\tau}_{c1})][u_c(t-{{\tau}_{c1}})-u_c(t-{{\tau}_{c2}})] 
\end{aligned}
\end{equation}
The control function is heuristic and open-loop in nature. Figure ~\ref{Fig:2}(b) provides an idea of the nature of this function. It has been tested in vast simulations over many vehicle parameters and in each case, suitable control function parameters could be found that yield trajectory restoration. 
Hence, it can be said that this heuristic function works for a large set of vehicles to restore their trajectory post-collision. The Results section includes two typical collision scenarios in simulation and studies the effect of the proposed controller to bring the vehicle back to its original path.
\section{Simulation Results}
The post-collision trajectory control using the above heuristic functions is applied to both the generalized Ackermann vehicle model and the 3DOF MATLAB single-track vehicle model (considered as the reference model). Table ~\ref{tab:1} specifies the vehicle parameters utilized for the post-collision trajectory restoration.
\begin{figure}[!htpb]
    \centering
    \begin{subfigure}{0.5\textwidth}
        \includegraphics[width=\linewidth]{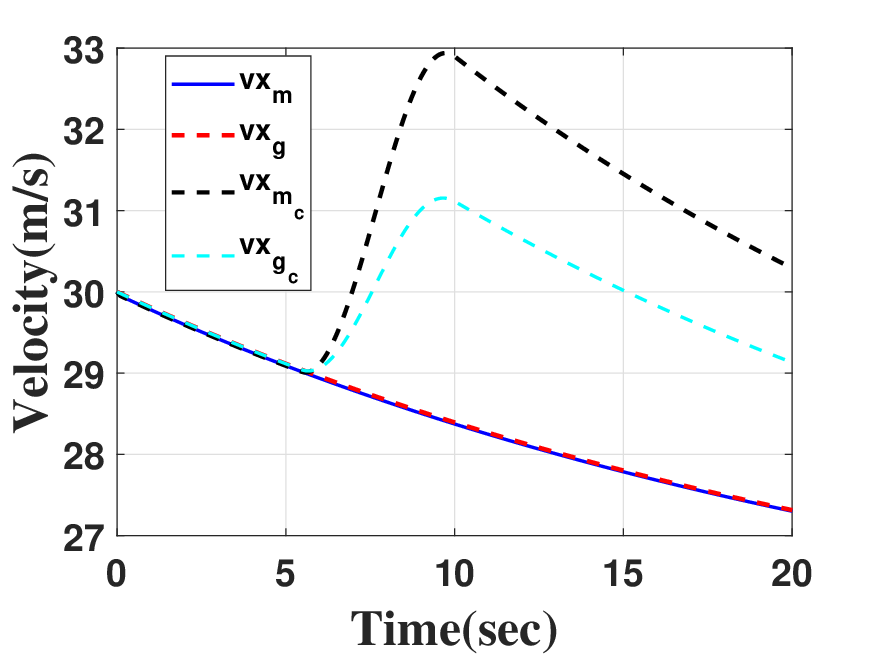}
        \caption{Longitudinal velocity ($v_x$)}
    \end{subfigure}\hfill
    \begin{subfigure}{0.5\textwidth}
        \includegraphics[width=\linewidth]{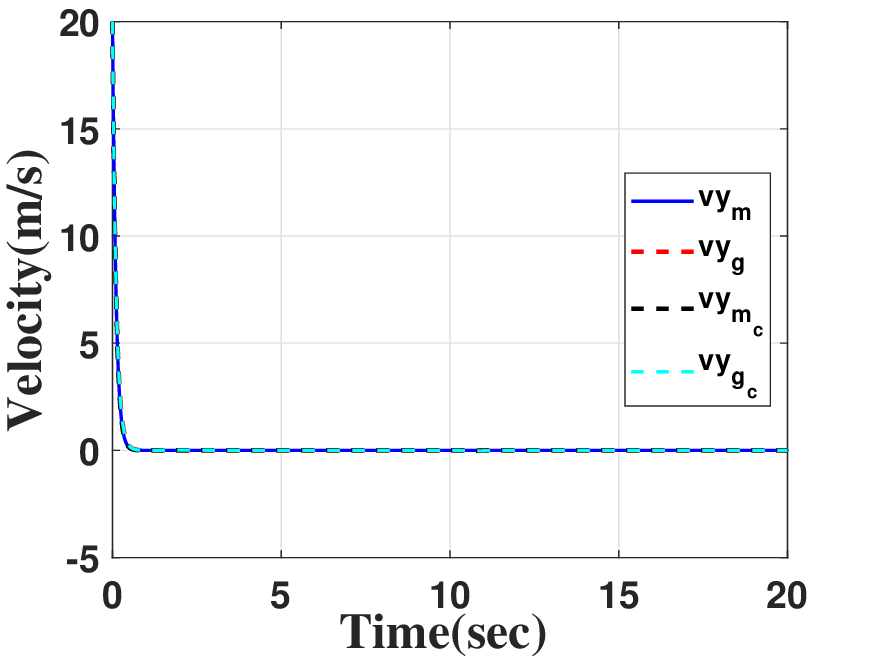}
        \caption{Lateral velocity ($v_y$)}
    \end{subfigure}

    \vspace{1em}

    \begin{subfigure}{0.5\textwidth}
        \includegraphics[width=\linewidth]{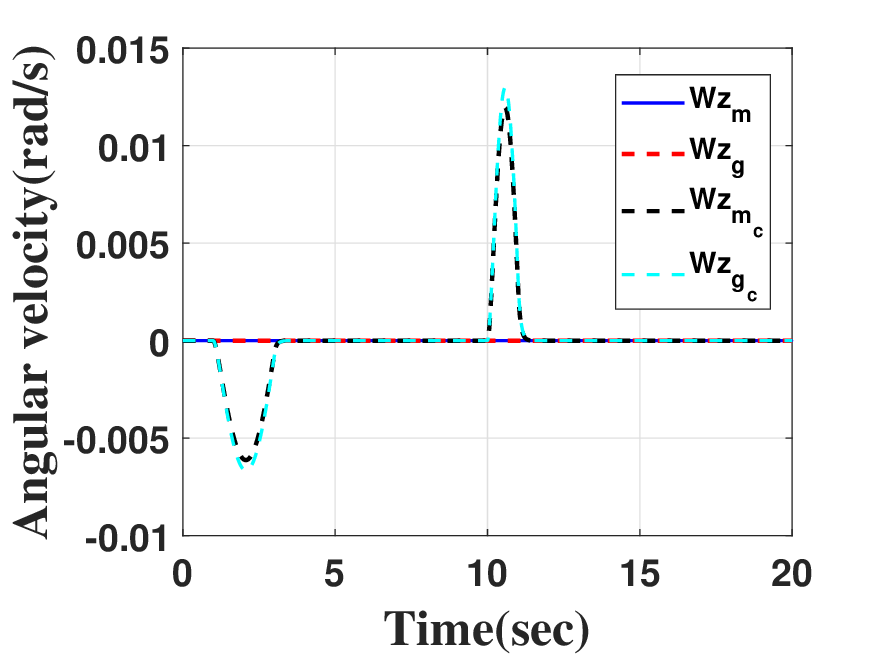}
        \caption{Angular velocity ($w_z$)}
    \end{subfigure}\hfill
    \begin{subfigure}{0.5\textwidth}
        \includegraphics[width=\linewidth]{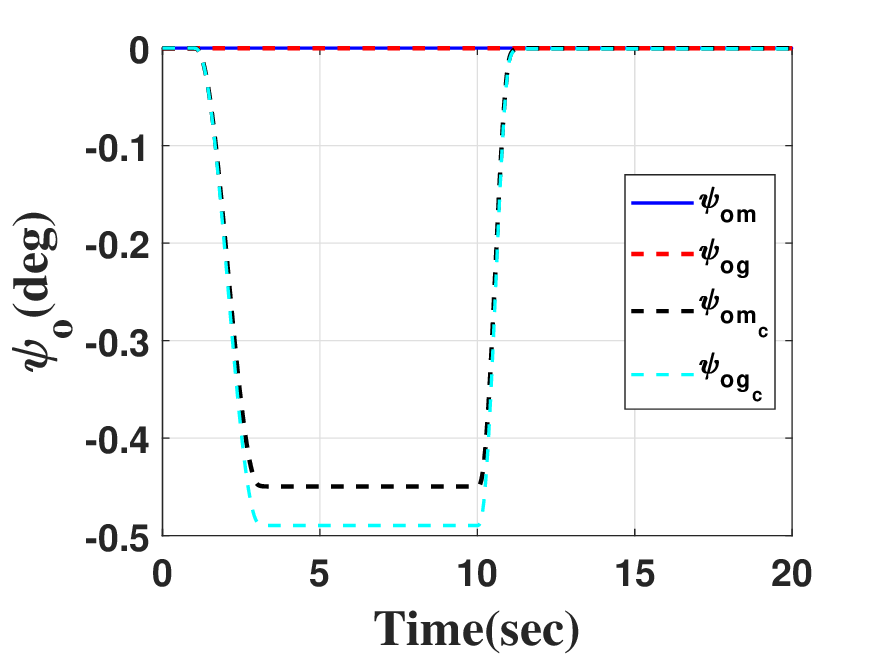}
        \caption{Orientation angle ($\psi_o$)}
    \end{subfigure}

    \vspace{1em}
    \caption{Case 1 collision scenario: (a) Longitudinal velocity ($v_x$), (b) Lateral velocity ($v_y$), (c) Angular velocity ($w_z$), (d) Orientation angle ($\psi_o$)}
    \label{Fig:3}
\end{figure}
The simulations validate the working of the proposed steering and tractive force function to restore the trajectory of a vehicle under the various post-collision scenarios
for both the generalized vehicle model and the reference model.
The collision is assumed to occur at the moment $t = 0$ to the lateral direction of the vehicle and the initial conditions in the lateral velocity and yaw angle emulate the collision condition. In the post-collision scenarios, recovery is considered for the original trajectory being a straight path only. So, the desired path of the vehicle is considered as $X(t)=f(v_x(t))$,$Y(t)=0$.
\par
Initially, the vehicle was traveling with the longitudinal velocity of $v_{x0}=30 \mathrm{m/s}$ at the time of the collision in both Case 1 and Case 2. The results associated with the MATLAB reference model without the control and with control are indicated by the suffix {'m'}, and {'m\textsubscript{c}'} respectively. 
 \begin{figure}[!htpb]
	\centerline{\includegraphics[width=0.6\columnwidth]{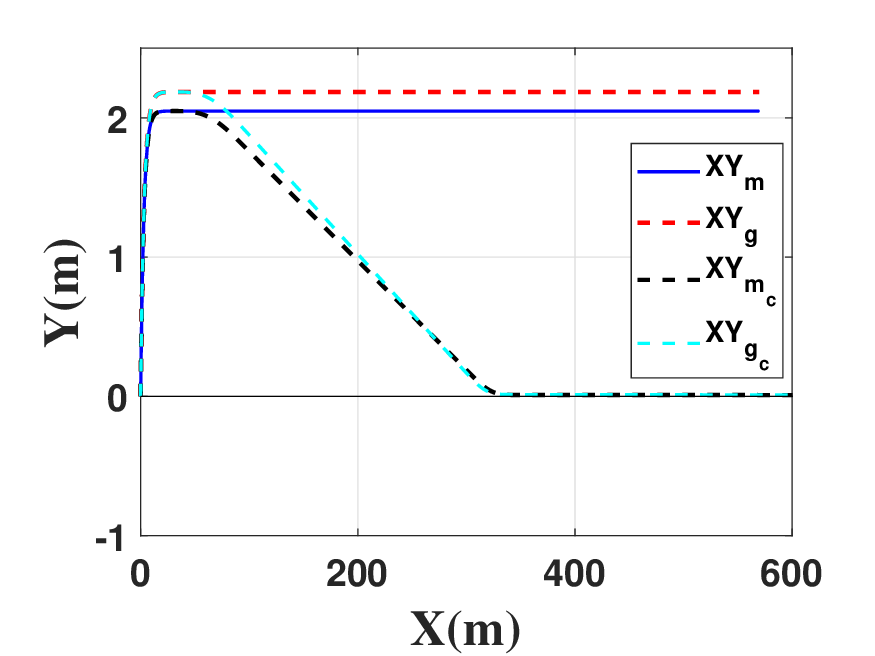}}
	\caption{Post-collision trajectory restoration for Case 1 collision scenario}
	\label{Fig:4}
\end{figure}
\par In the generalized Ackermann model, the suffixes {'g'}, and {'g\textsubscript{c}'} are used to indicate the plots associated without the control and with the control. Table ~\ref{tab:2} shows the parameters of the steering and the tractive force function used for the trajectory restoration of the vehicle model in the post-collision scenarios.
Due to the inclusion of the complex nonlinear interaction terms associated with the steering angle in the generalized model, different control parameter values are required for the generalized model and the reference model to restore the trajectory in the post-collision scenarios. This also indicates the versatility of the control function to work on various models of vehicle dynamics for post-collision trajectory restoration.
\begin{table}[!htpb]
\centering
	\caption{\centering{Vehicle Model Parameters}}
	\setlength{\tabcolsep}{8pt}
	\begin{tabular}{|p{30 pt}|p{160 pt}|p{110 pt}|}
		\hline
		Symbol& 
		Parameters& 
		Value \\
		\hline
		$m $& Vehicle Mass& $1750 \mathrm{Kg}$\\
		\hline
		$\mu_0$& Frictional constant& $0.015$ \\
		\hline
		$\mu_1$& Frictional coefficient& $7 \times 10^-6\mathrm{{s^2}{m^-2}}$\\
		\hline
		$g$& Gravitational acceleration& $9.8 \mathrm{m/s^2}$ \\
		\hline
		$I_z$& Moment of Inertia w.r.t to z axis& $2350\mathrm{Kgm^2}$ \\
		\hline
            $C_{{\alpha}{f}}$& Front tire cornering stiffness& $12\times10^4 \mathrm{N/rad}$ \\
		\hline
		$C_{{\alpha}{r}}$& Rear tire cornering stiffness& 
            $12\times10^4\mathrm{N/rad}$ \\
		\hline
		$K_d$& Aerodynamics Drag coefficient& $0.98\mathrm{N{{m}^-2}{s^-2}}$\\
		\hline
	\end{tabular}
	\label{tab:1}
\end{table}
\subsection{Case 1 (Lateral Collision at the C.G. of the vehicle):}
The collision happens in the lateral direction of the vehicle due to which the vehicle is driven laterally (i.e., towards the y-axis) due to the collision. The collision is emulated by providing the initial lateral velocity of $v_{y0}=20\mathrm{m/s}$ and initial angular velocity of $w_{z0}=0 \mathrm{rad/s}$ at and about the C.G. at $t=0\mathrm{sec}$. 
Figure~\ref{Fig:3}(a), ~\ref{Fig:3}(b), ~\ref{Fig:3}(c), and ~\ref{Fig:3}(d) show the longitudinal, lateral, angular velocities, and orientation angle of the generalized vehicle model and the reference model during the collision with control and without control.
From Figure ~\ref{Fig:4}, it can be observed that the trajectory is restored after the collision for both the generalized vehicle model and the reference model when the control function is applied. Whereas, due to the collision the vehicle is distracted from the track and moving parallel to the track if no control function is applied.
\begin{table*}[!h]
\centering
	\caption{\centering{Steering and Tractive force function Parameters}}
	\setlength{\tabcolsep}{8pt}
	\begin{tabular}{|p{25 pt}|p{195 pt}|p{195 pt}|}
		\hline
		Cases& 
		Proposed Model& 
		Reference Model\\
		\hline
		Case 1&$A_1=0.175$, $K_1=-1.91$, $A_c=900\mathrm{N}$, $K_{dir}=-0.2$, $\tau_0=1\mathrm{s}$, $\tau_1=3\mathrm{s}$, $\tau_2=10\mathrm{s}$, $\tau_3=11\mathrm{s}$,$\tau_{c1}=5.443\mathrm{s}$, $\tau_{c2}=10\mathrm{s}$&$A_1=0.175$, $K_1=-1.818$, $A_c=1500\mathrm{N}$, $K_{dir}=-0.2$, $\tau_0=1\mathrm{s}$, $\tau_1=3\mathrm{s}$, $\tau_2=10\mathrm{s}$, $\tau_3=11\mathrm{s}$,$\tau_{c1}=5.443\mathrm{s}$, $\tau_{c2}=10\mathrm{s}$\\
		\hline
		Case 2&$A_1=0.175$, $K_1=-1.4665$, $A_c=900\mathrm{N}$, $K_{dir}=-0.5$, $\tau_0=1\mathrm{s}$, $\tau_1=5.195\mathrm{s}$, $\tau_2=10\mathrm{s}$, $\tau_3=11\mathrm{s}$,$\tau_{c1}=5.443\mathrm{s}$, $\tau_{c2}=10\mathrm{s}$&$A_1=0.175$, $K_1=-1.353$, $A_c=1550\mathrm{N}$, $K_{dir}=-0.5$, $\tau_0=1\mathrm{s}$, $\tau_1=5.195\mathrm{s}$, $\tau_2=10\mathrm{s}$, $\tau_3=11\mathrm{s}$,$\tau_{c1}=5.443\mathrm{s}$, $\tau_{c2}=10\mathrm{s}$\\
		\hline
	\end{tabular}
	\label{tab:2}
\end{table*}
\subsection{Case 2 (Lateral Collision not at the C.G. of the vehicle):}
\begin{figure}[!htpb]
    \centering
    \begin{subfigure}{0.5\textwidth}
        \includegraphics[width=\linewidth]{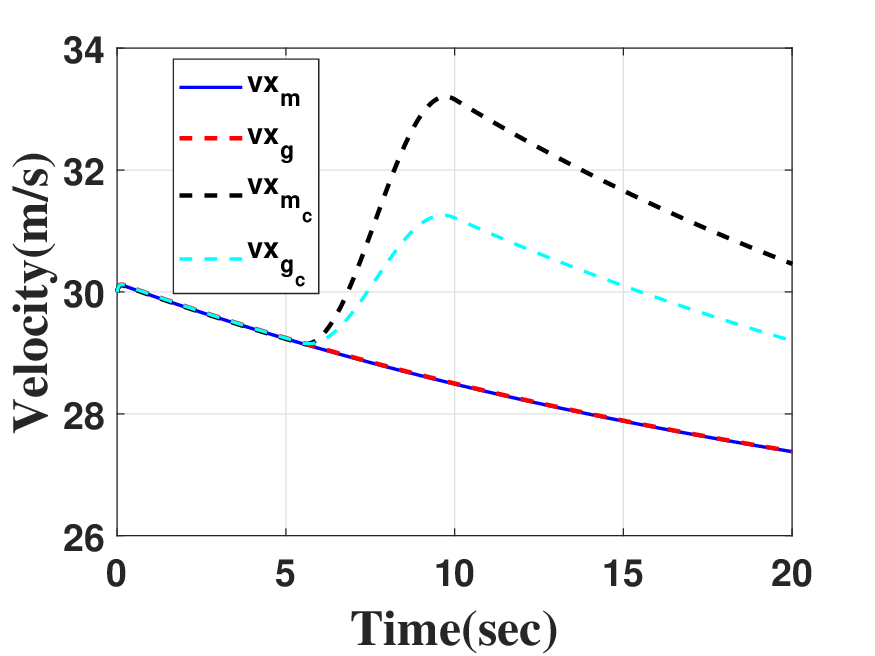}
        \caption{Longitudinal velocity ($v_x$)}
    \end{subfigure}\hfill
    \begin{subfigure}{0.5\textwidth}
        \includegraphics[width=\linewidth]{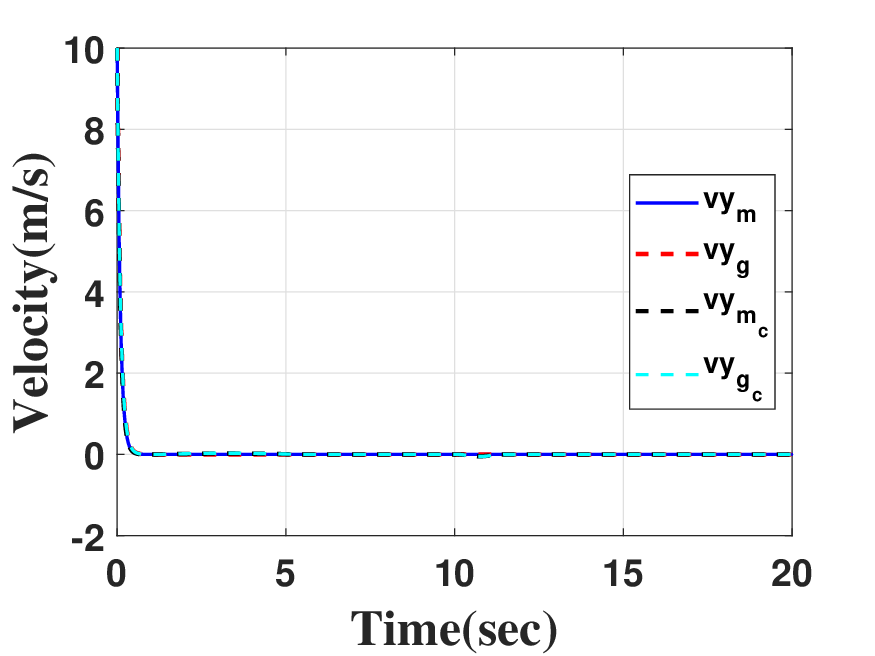}
        \caption{Lateral velocity ($v_y$)}
    \end{subfigure}

    \vspace{1em}

    \begin{subfigure}{0.5\textwidth}
        \includegraphics[width=\linewidth]{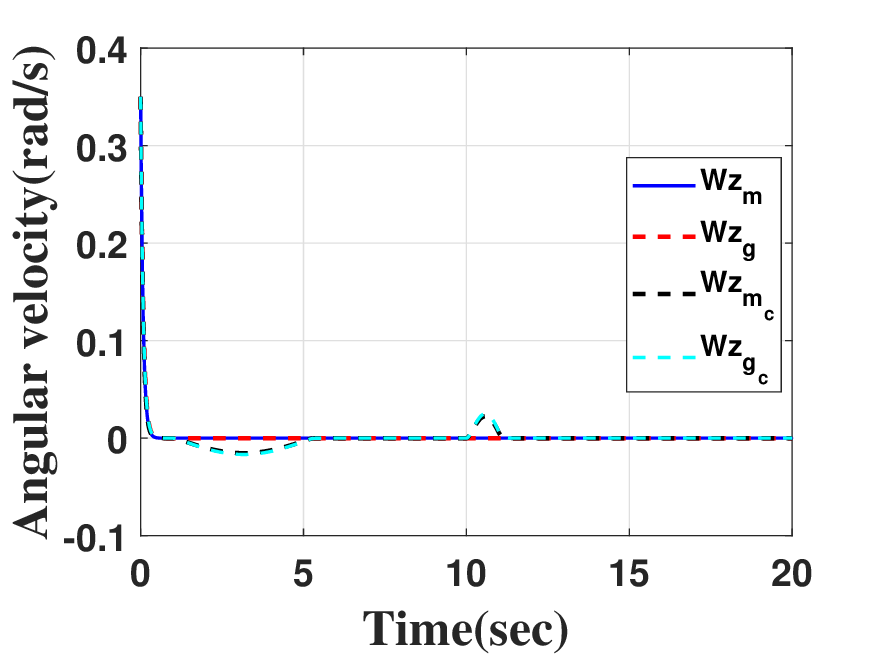}
        \caption{Angular velocity ($w_z$)}
    \end{subfigure}\hfill
    \begin{subfigure}{0.5\textwidth}
        \includegraphics[width=\linewidth]{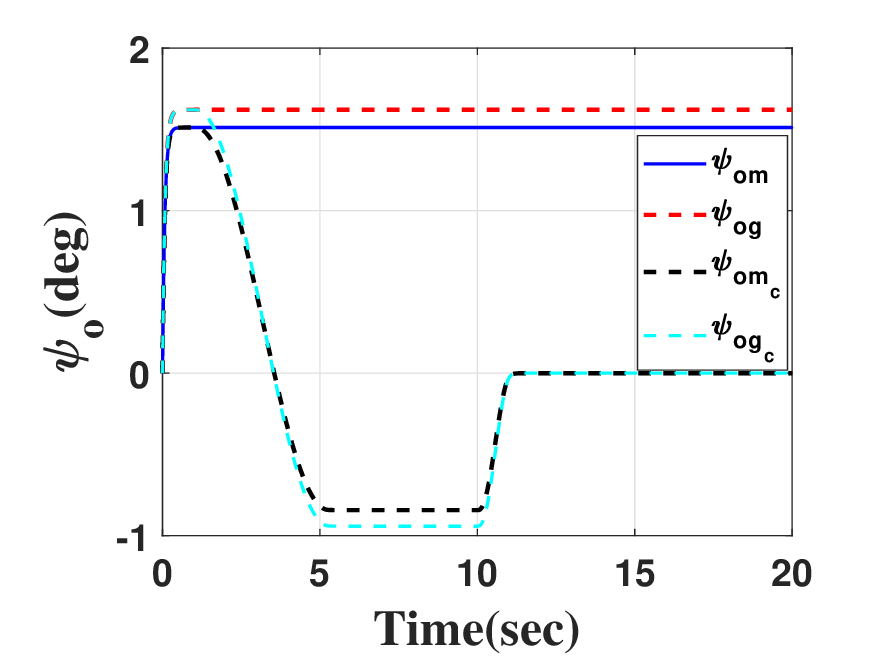}
        \caption{Orientation angle ($\psi_o$)}
    \end{subfigure}

    \vspace{1em}
    \caption{Case 2 collision scenario: (a) Longitudinal velocity ($v_x$), (b) Lateral velocity ($v_y$), (c) Angular velocity ($w_z$), (d) Orientation angle ($\psi_o$)}
    \label{Fig:5}
\end{figure}

\begin{figure}[!htpb]
    \centerline{\includegraphics[width=0.6\columnwidth]{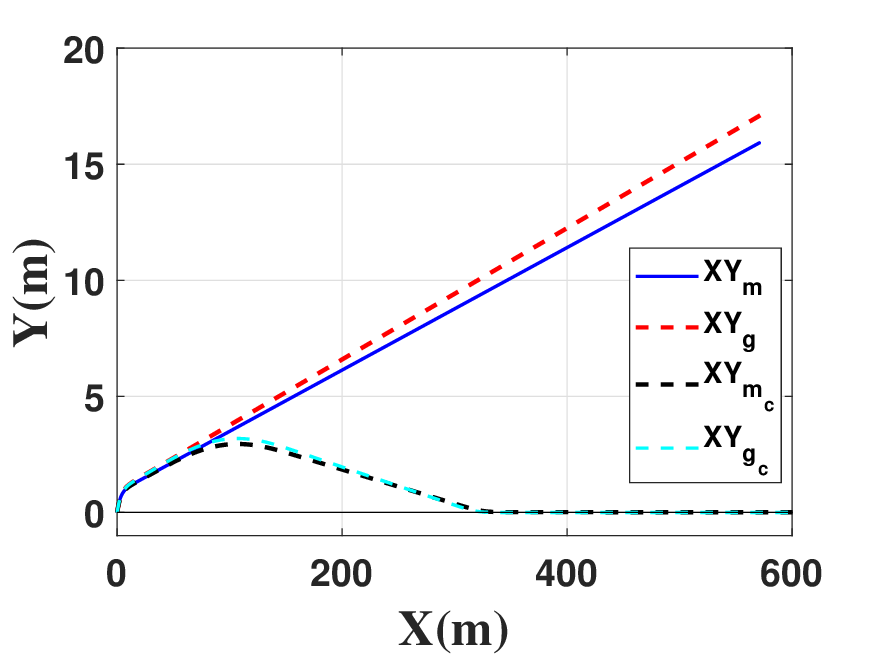}}
	\caption{Post-collision trajectory restoration for Case 2 collision scenario}
	\label{Fig:6}
\end{figure}

The collision happens in the lateral direction of the vehicle due to which the vehicle is driven laterally (i.e., towards the y-axis), and its orientation is also altered. 
The collision is emulated by providing the initial angular velocity of $w_{z0}=0.35 \mathrm{rad/s}$ and initial lateral velocity of $v_{y0}=10\mathrm{m/s}$at and about the C.G. at $t=0\mathrm{sec}$.
Figure ~\ref{Fig:5}(a), ~\ref{Fig:5}(b), ~\ref{Fig:5}(c), and ~\ref{Fig:5}(d) show the longitudinal, lateral, angular velocities, and orientation angle of the generalized vehicle model and the reference model during the collision with control and without control.
From Figure ~\ref{Fig:6}, it can be observed that the trajectory is restored after the collision for both the generalized vehicle model and the reference model when the control function is applied. Whereas, due to the collision the vehicle is distracted from the track if no control function is applied.

\section{Discussion}
The proposed steering and tractive-force shaping strategy achieves post-collision restoration by (i) attenuating yaw-rate and lateral-velocity transients immediately after the post-impact state initialization, and (ii) progressively re-aligning the vehicle heading and lateral position with the desired path while recovering forward motion. A key practical implication is that longitudinal-speed variations during recovery couple into the lateral--yaw dynamics and can alter the effective control authority; consequently, constant-$v_x$ assumptions may lead to inaccurate predictions of the transient response in post-collision scenarios. In the presented simulations, the combined steering--traction action reduces peak yaw response and lateral deviation compared with steering-only recovery, indicating that longitudinal actuation can materially influence lateral stabilization when nonlinear coupling terms are retained. The recovery characteristics (e.g., overshoot, oscillation, and convergence rate) depend on controller parameterization and the severity of the post-impact initial conditions, motivating systematic robustness testing and constraint-aware implementations.

\section{Reproducibility and Implementation Notes}
All simulations are performed using a fixed integration step $\Delta t$ (report value). The collision event is modeled through post-impact initialization of the dynamic states; specifically, representative post-collision values of $v_x$, $v_y$, $r$, and the path-relative errors (e.g., lateral error and heading error) are used as initial conditions (report either the discrete set of cases or the parameter ranges). Recovery performance is quantified using metrics such as time-to-recovery, peak lateral deviation, and peak yaw-rate and/or sideslip, with explicit thresholds defining successful restoration (report thresholds). For transparency and repeatability, the following items should be reported or tabulated: vehicle parameters (mass, wheelbase, $C_f$, $C_r$, yaw inertia), tire/road assumptions (friction coefficient and tire model), actuator limits (steering angle/rate, tractive/braking limits, if enforced), controller parameters, and the complete set of post-impact initialization cases.

\section{Limitations and Future Work}
This study focuses on post-collision recovery control and simulation-based evaluation using single-track vehicle models. The collision is not modeled as a contact/impact process; instead, it is represented by post-impact state initialization. Therefore, the proposed method is intended for the recovery phase and does not predict impact forces, deformation, or the state transition caused by contact. Moreover, while the generalized model retains nonlinear coupling terms and varying longitudinal speed, the current evaluation does not yet provide systematic robustness characterization with respect to uncertainty in key parameters (e.g., tire cornering stiffness, friction coefficient, mass and CG shift), nor does it comprehensively analyze constraint handling under actuator saturation and combined-slip tire-force limits (clarify which constraints are included, if any). Future work will (i) incorporate combined-slip tire constraints and explicit steering/traction actuator limits within the recovery strategy, (ii) conduct Monte-Carlo studies over impact conditions, road friction, and vehicle-parameter variations to quantify success rate and worst-case behavior, and (iii) validate closed-loop recovery in higher-fidelity simulation and/or experimental platforms.

\section{CONCLUSIONS}
Post-collision trajectory restoration using heuristic steering and tractive force function is undertaken in this work. The generalized single-track Ackermann vehicle model from \cite{ghosh2023collision} is taken that considers the frictional force and other forces in the longitudinal, lateral, and yaw dynamics, along with complex nonlinear interaction terms due to the steering angle. The post-collision trajectory restoration is also studied for the 3DOF MATLAB single-track vehicle model. 
Based on the simulations, it can be inferred that the proposed heuristic control function can restore the trajectory of both the reference model and the generalized Ackermann vehicle model. It can be noted that post-collision the vehicle continues to drive with high speed and does not apply complete brakes but recovers to its original path using the proposed heuristic control. This proposed heuristic control function may be used to develop an efficient automated driving algorithm to deal with unforeseen collision encounters of the vehicles by strategically using it along with feedback control architecture in future research.

\bibliographystyle{unsrtnat}
\bibliography{Refrence1}  






\end{document}